\documentclass[11pt]{article}

\usepackage[preprint]{acl}

\usepackage{times}
\usepackage{latexsym}
\usepackage{amsmath}
\usepackage{booktabs}
\usepackage[capitalise]{cleveref}
\crefname{figure}{Figure}{Figures}
\Crefname{figure}{Figure}{Figures}

\usepackage[T1]{fontenc}

\usepackage[utf8]{inputenc}

\usepackage{microtype}

\usepackage{inconsolata}

\usepackage{graphicx}

\title{BioBigBird: A Sparse Attention Model for Long-Range Dependency Processing in Biomedical Text}

\author{
  \textbf{Roshan Balaji},
  \textbf{Pavan Kumar S},
  \textbf{Vasudev Gupta},
  \textbf{Sreejith N},
  \textbf{Keerthana Sridhar},
  \textbf{Nirav Bhatt}
  \\
  Wadhwani School of Data Science and AI, \\
  Indian Institute of Technology Madras, Chennai, India
}

\begin{document}
\maketitle
\begin{abstract}
While domain-specific Large Language Models (LLMs) have encoded vast biomedical knowledge, their limited context windows often hinder a deep understanding of nuanced relationships within and across texts. To address this limitation, we introduce BioBigBird, a bidirectional language model pre-trained on extensive biomedical literature and clinical data, specifically designed to handle long-range dependencies. BioBigBird leverages a sparse attention mechanism to process sequences up to 4096 tokens, and its training incorporates a multi-stage process to mitigate noise from the large-scale pre-training corpus. We further enhance its performance by employing a multi-task learning (MTL) framework that jointly optimizes for Named Entity Recognition and Relation Extraction. Comprehensive evaluations on the BLURB benchmark reveal that our MTL-enhanced BioBigBird achieves highly competitive results against state-of-the-art models. Our work contributes an effective methodology for developing powerful, long-context language models for specialized domains, demonstrating the value of extended sequence processing for complex text analysis. Our models are publicly available at \url{https://huggingface.co/collections/bisectgroup/biobigbird}.
\end{abstract}

\section{Introduction}

The field of biomedicine is characterised by a massive and ever-growing volume of unstructured text, from millions of publications in repositories like PubMed \citep{pubmed2011pubmed} to detailed clinical narratives in electronic health records \citep{johnson2016mimic}. Effectively harnessing this data requires advanced text-mining tools. Core to this endeavour are the tasks of Biomedical Named Entity Recognition (BioNER) and Relation Extraction (BioRE), which automatically identify key entities (e.g., genes, diseases, drugs) and the relationships between them. These tasks are foundational for building structured knowledge bases, accelerating research, and discovering novel insights from the vast body of literature \citep{wang2023pre}.

The advent of the Transformer architecture \citep{vaswani} and pre-trained language models like BERT \citep{devlin} has revolutionised Natural Language Processing (NLP). By learning deep bidirectional representations of text, these models achieve state-of-the-art performance on a wide range of tasks after fine-tuning. However, their success comes with a critical limitation: the self-attention mechanism at their core has a computational and memory complexity that is quadratic to the input sequence length ($O(n^2)$). The sequence complexity restricts models like BERT to short contexts, typically 512 tokens. Consequently, input must be truncated or segmented when applied to full-text articles or lengthy clinical summaries, leading to a loss of long-range contextual information and suboptimal performance on downstream tasks.

Two key research directions have emerged to address these challenges. First, work by \citet{yugu} has demonstrated that pre-training language models from scratch on domain-specific corpora (e.g., biomedical text) significantly outperforms continually training a general-domain model. Their introduction of the BLURB benchmark has been instrumental in tracking progress in biomedical language understanding. Second, models like BigBird \citep{zaheer} have been proposed to overcome the sequence length limitation. BigBird utilises a sparse attention mechanism that reduces the complexity from quadratic to linear, enabling the processing of sequences up to 4096 tokens or longer, making it possible to analyse entire documents without truncation.

Building on these advances, this paper introduces \textbf{BioBigBird}, a biomedical language model designed to address the challenges of domain-specificity and long-range context holistically. Our approach makes two primary contributions. First, we employ a multi-stage \textbf{curriculum learning} strategy for pre-training. We begin training with shorter sequences (512 tokens) to efficiently learn fundamental linguistic patterns and stabilise the model, before graduating to longer sequences (up to 4096 tokens) to capture complex, long-range dependencies. This staged approach, which preserves optimiser states across stages, is more computationally efficient and effective than training on long sequences from the outset.

Second, to maximise performance on biomedical tasks, we integrate a \textbf{multi-task learning (MTL)} framework for fine-tuning. As different BioNER and BioRE datasets are often annotated for distinct entity and relation types, MTL allows a single model to learn from multiple datasets simultaneously. MTL encourages the model to learn shared, generalisable representations, improving performance and data efficiency, a principle shown to be effective in prior work \citep{crichton2017code, 10.1093/bioinformatics/bty869}.

In summary, our contributions are:
\begin{itemize}
    \item The development and public release of BioBigBird,\footnote{\url{https://huggingface.co/collections/bisectgroup/biobigbird}} a new long-sequence language model pre-trained from scratch on a comprehensive biomedical and clinical text corpus.
    \item A demonstration of an effective and efficient multi-stage curriculum for pre-training long-context models in a specialised domain.
    \item The integration of a multi-task learning framework to synergistically improve performance on diverse BioNER and BioRE tasks.
    \item A comprehensive evaluation on the BLURB benchmark, showing that BioBigBird achieves highly competitive results compared to existing state-of-the-art models.
\end{itemize}

\section{Background and Related Work}

\subsection{Domain-Specific Pre-Training}

In natural language processing, domain-specific pre-training is an approach to improve the performance of models within specialized domains \citep{yugu}. Our research focused on the biomedical field, where we aimed to optimize the model's understanding of complex language structures and domain-specific terminologies. To achieve this, we utilized biomedical corpora. The resulting model is expected to perform better and be more applicable to tasks within the biomedical domain. Our study showcases the effectiveness of this targeted approach in improving the model's domain-specific capabilities.

\subsection{BigBird Attention Mechanism}

The attention mechanism calculates the attention weights by applying a softmax function to the scaled dot-product of the query and key matrices. These attention weights are multiplied by the value matrix to obtain the final output.

In this work, we employed the BigBird architecture, a transformer-based model known for efficiently handling longer sequences. The architecture's key feature is its ability to capture long-range dependencies by processing the input sequence in chunks, thereby reducing the computational complexity of the model. Our BigBird model consisted of 12 transformer layers, each with 768 hidden units and 12 attention heads, resulting in ${\approx}$113M trainable parameters. This design ensured that our model could capture complex patterns in biomedical literature, which often span long sequences and require processing vast amounts of data. The selection of the BigBird architecture was a crucial factor in developing a high-performing domain-specific language model.

The attention mechanism is described as follows:
\begin{equation}
    \mathrm{Attention}(\mathbf{Q},\mathbf{K},\mathbf{V}) = \mathrm{softmax}\left(\frac{\mathbf{Q}\mathbf{K}^\top}{\sqrt{d_k}}\right)\mathbf{V}
\end{equation}
where $\mathbf{Q}$, $\mathbf{K}$, and $\mathbf{V}$ are the query, key, and value matrices, consisting of queries and keys of dimension $d_k$ and values of dimension $d_v$. Each element in $\mathbf{Q}$ is mapped to an element in $\mathbf{K}$ with a value in $\mathbf{V}$.

\begin{figure}[t]
    \centering
    \includegraphics[width=\linewidth]{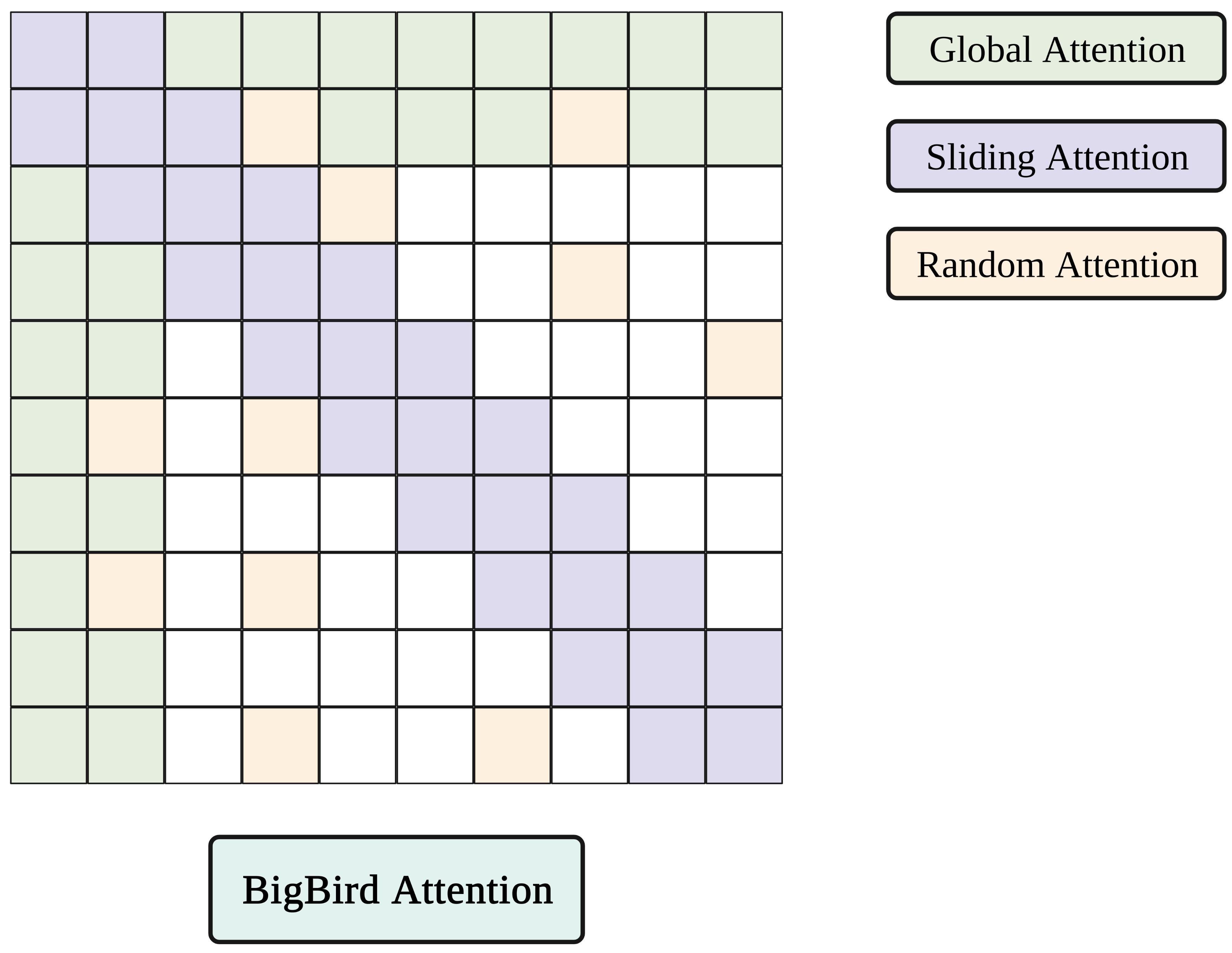}
    \caption{Attention mechanism used in BigBird. The white colour indicates the absence of attention. Random, global, and sliding connections are combined to compute the attention scores for longer sequences efficiently.}
    \label{fig:attn}
\end{figure}

The key and value matrices are selected based on the global, sliding, and random connections as shown in \cref{fig:attn}.

We explain the BigBird architecture in the following sections.

\subsubsection{Global Attention}
In global attention, each query attends to all the other tokens in the sequence and is attended to by all the tokens. \Cref{fig:attn} corresponds to the global attention mechanism, wherein the first, second, and last tokens are considered global tokens. The computational complexity is $O(n)$.

\subsubsection{Sliding Attention}
In sliding attention, each query attends to its neighbouring tokens. This can be implemented by copying the sequence of key tokens two times and shifting every element to the right in one of the copies and to the left in the other. The final attention score can be computed by multiplying query sequence vectors by these three sequences. The computational complexity is $O(n)$.

\subsubsection{Random Attention}
In random attention, each query token attends to a few random tokens. This means the model randomly gathers some tokens and computes their attention score. The computational complexity is $O(n)$.

\subsubsection{Graph Theory View}

\begin{figure}[t]
    \centering
    \includegraphics[width=\linewidth]{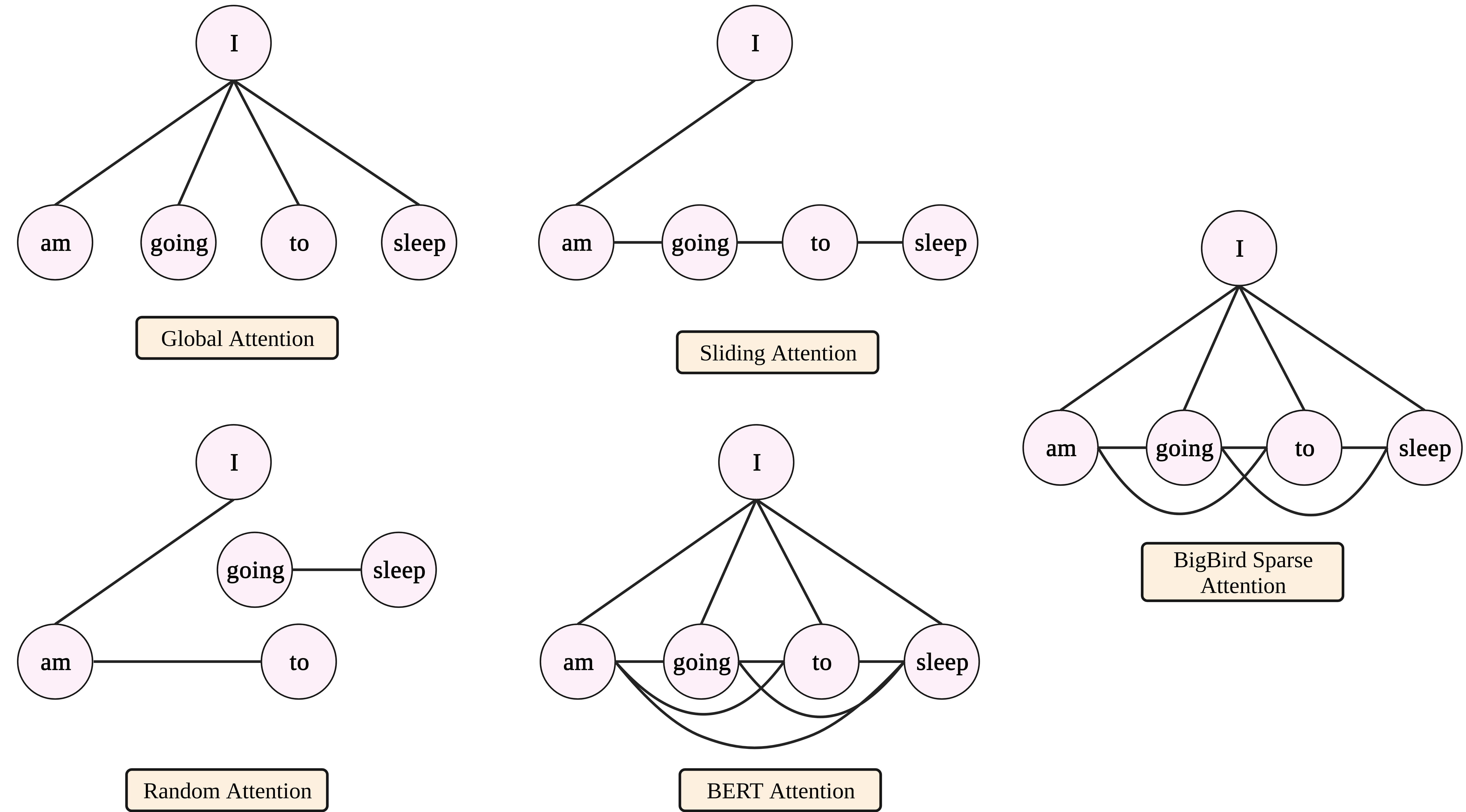}
    \caption{The comparison of BERT and BigBird attention mechanisms.}
    \label{fig:graph}
\end{figure}

BigBird attention can be understood with graph theory. \Cref{fig:graph} shows global, sliding, and random connections in the graph theory. Each node corresponds to a token, and each line represents the attention score. If no connection is made between two tokens, then the attention score is assumed to be 0.

When the model needs to share information between two nodes (or tokens), information can travel across various other nodes in the path since all the nodes are not directly connected in a single layer. For example, suppose the model needs to associate `going' with `sleep'. In that case, if only sliding attention is present, the flow of information among those two tokens can be defined by the path: going $\rightarrow$ to $\rightarrow$ sleep (i.e., it will have to travel over one other token). Hence, we may need multiple layers to capture the entire sequence of information. BERT attention can capture this in a single layer. If we introduce some global tokens, information can travel via the path: going $\rightarrow$ I $\rightarrow$ sleep (which is shorter). If we introduce random connections, it can travel via going $\rightarrow$ sleep. With the help of random connections and global connections, information can travel rapidly (with just a few layers) from one token to the next.

If we have many global tokens, we may not need random connections since there will be multiple short paths through which information can travel.

\subsection{Masked Language Modelling}

We utilized a Masked Language Modeling (MLM) approach, as described by \citet{devlin}, to pre-train the BioBigBird model. This approach helps the model understand a language's syntactic structures better. MLM is a fundamental technique used in natural language processing. It involves masking a percentage of tokens in a given text and asking the model to predict the masked tokens based on contextual information from the surrounding words. The MLM objective enables the representation to fuse the left and the right context, which allows us to pre-train a deep bidirectional Transformer. This contributes to developing a robust and context-aware language model for improved performance in downstream natural language processing tasks.

\begin{figure*}[t]
    \centering
    \includegraphics[width=\textwidth]{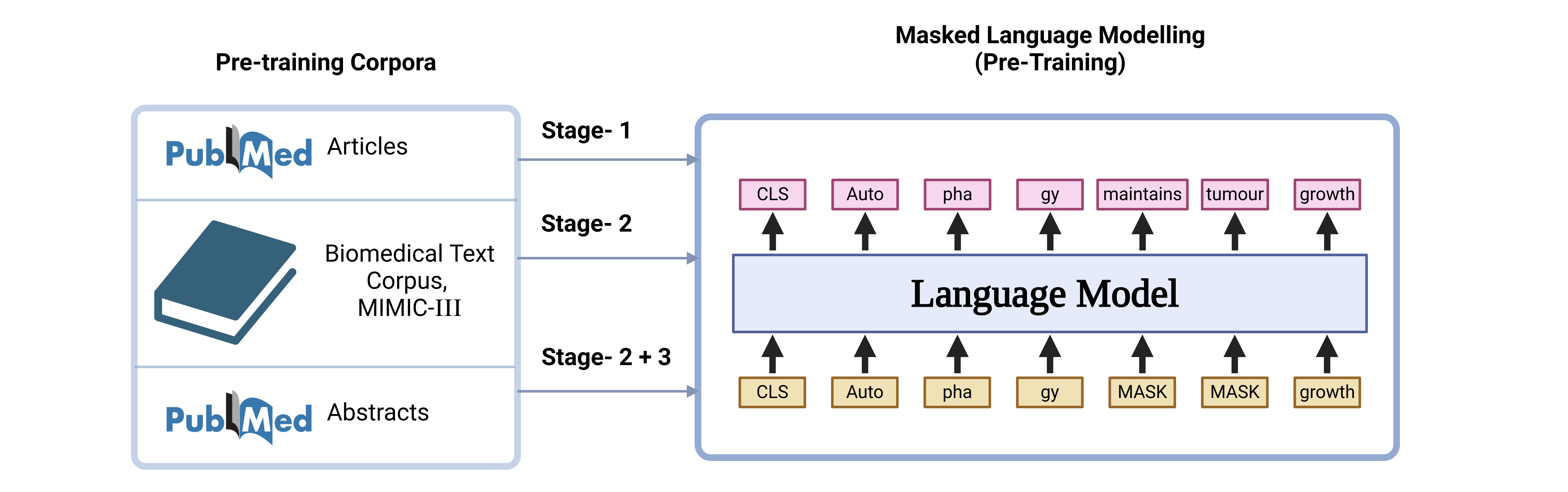}
    \caption{Overview of our approach BioBigBird: We collect documents related to the biomedical corpora and use a three-stage approach to pre-train the language model using masked language modelling.}
    \label{fig:overview}
\end{figure*}

\section{Methodology}

\subsection{Pre-training Corpus}
The datasets used to pre-train BioBigBird are from PubMed and MIMIC-III. PubMed is a database of abstracts and references in the biomedical field. All the abstracts available in the database till 2021 are collected for model pre-training. The freely available articles are also collected for the same. MIMIC-III is a freely available critical care database. It contains de-identified electronic health records of over 40,000 patients who were admitted to the intensive care units of Beth Israel Deaconess Medical Center between 2001 and 2012. The statistics of the datasets are shown in \cref{tab:data_stats,tab:article_stats}. We applied a rigorous pre-processing pipeline to the PubMed articles corpus. To ensure content quality, we retained only articles that contained both an introduction and a conclusion section and had a minimum length of 512 tokens. We further refined the dataset by removing duplicate articles, figures, tables, and any lines where more than 70\% of the characters were numeric. For the final training data, we used only the textual content from the introduction and conclusion sections, truncating any documents longer than 4096 tokens.

\begin{table}[t]
    \centering
    \small
    \setlength{\tabcolsep}{4pt}
    \caption{Statistics of the pre-training datasets.}
    \label{tab:data_stats}
    \begin{tabular}{lccc}
        \toprule
        \textbf{Dataset} & \textbf{Disk Space} & \textbf{\# Samples} & \textbf{\# Tokens} \\
        \midrule
        PubMed Articles  & 80 GB  & 2.3 M & 10 B \\
        PubMed Abstracts & 31 GB  & 42 M  & 10 B \\
        MIMIC-III        & 3.5 GB & 2 M   & 2 B  \\
        \bottomrule
    \end{tabular}
\end{table}

\begin{table}[t]
    \centering
    \small
    \caption{Statistics of the PubMed articles before and after pre-processing.}
    \label{tab:article_stats}
    \begin{tabular}{lcc}
        \toprule
        \textbf{Dataset} & \textbf{Disk Space} & \textbf{\# Samples} \\
        \midrule
        Raw Articles           & 300 GB & 4.7 M \\
        Pre-processed Articles & 80 GB  & 2.3 M \\
        \bottomrule
    \end{tabular}
\end{table}

\subsection{Multi-Stage Pre-training Strategy}

Pre-training a language model on a large corpus of text is crucial for effectively performing downstream tasks. In this study, we pre-trained our model on an extensive corpus of the biomedical literature, including PubMed full articles and clinical data such as MIMIC-III, using a multi-stage pre-training approach as shown in \cref{fig:overview}. Our model underwent three stages of pre-training, with each stage designed to enhance the model's ability to capture complex patterns in biomedical language. In stage 1, we pre-trained on PubMed full articles up to a sequence length of 4096, enabling our model to learn long-range dependencies in the biomedical literature. In stage 2, we continued pre-training the model from stage 1 on MIMIC-III and PubMed abstracts up to a sequence length of 1024, which allowed our model to capture variations in language and terminology across different domains. Finally, in stage 3, we continued pre-training the model from stage 2 on PubMed abstracts up to a sequence length of 512, further refining the model's ability to understand the language of biomedical research articles. Each pre-training stage was carried out for a specific number of steps, with stage 1 taking 1.89M steps, stage 2 taking 1.37M steps, and stage 3 taking 1.5M steps. This multi-stage pre-training approach equipped our model with a deep understanding of biomedical literature's complex structures and relationships, making it well-suited for downstream biomedical language understanding tasks.

\subsection{Fine-tuning Strategies}
The Named Entity Recognition (NER) problem involves analyzing a sentence with multiple tokens $[t_1, t_2, \ldots, t_n]$ and predicting whether each token belongs to the categories B, I, or O. Here, B signifies the beginning of an entity, I indicates being inside an entity, and O denotes being outside any entity. The goal is to assign these labels to each token.

The relation extraction (RE) problem is a task that identifies and classifies relationships between entities mentioned in the text. The nature of these relationships is determined by the specific entities mentioned within the text.

\subsubsection{Single-Task Fine-tuning}

A common strategy for fine-tuning models for NER and RE involves a single-task approach. In this method, separate models are fine-tuned for each specific entity type, as annotated datasets usually have only one or a few types of entities mentioned. The single-task model has a straightforward architecture, where the output from a pre-trained model is fed into a classification output layer. In the context of RE, the output from the pre-trained model is in the form of pooled output. On the other hand, for NER, the output is at the token level. The following are the definitions of loss functions for NER and RE in single-task settings.

Let $D_{\mathrm{NER}}$ be a NER dataset with $n$ training texts, each having length $L$. The loss function $L_{\mathrm{NER}}$ is defined as follows:
\begin{gather*}
    D_{\mathrm{NER}} = \{(\mathit{token}_i, y_i)_j\}, \quad \forall j \in \{1, \ldots, n\} \\
    L_{\mathrm{NER}} = -\sum_{j=1}^{n} \sum_{i=1}^{L} \log P(\hat{y}_i)
\end{gather*}
where $P(\hat{y}_i)$ represents the model's predicted probability for the correct class $y_i$ of each token.

Let $D_{\mathrm{RE}}$ be a RE dataset with $n$ training texts. The loss function $L_{\mathrm{RE}}$ is defined as follows:
\begin{gather*}
    D_{\mathrm{RE}} = \{(\mathit{text}, y)_j\}, \quad \forall j \in \{1, \ldots, n\} \\
    L_{\mathrm{RE}} = -\sum_{i=1}^{n} \log P(\hat{y}_i)
\end{gather*}
where $P(\hat{y}_i)$ represents the probability assigned by the model to the correct class $y_i$ for each text during classification.

For each task, separate models are trained using appropriate loss functions.

\subsubsection{Multi-task Fine-tuning}

This research addresses NER and RE challenges by proposing a multi-task learning framework that simultaneously tackles both tasks. The proposed method extends the multi-task learning paradigm to the sub-tasks within NER and RE. For example, some of the various tasks involved are disease entity recognition, chemical entity recognition, drug-drug relation extraction, chemical-disease relation extraction, etc. The primary goal is to enhance the overall performance of NER and RE by leveraging shared representation learning.

An important characteristic of the BioNER and BioRE tasks is the limited availability of supervised training data. We propose a multi-task learning approach to address this problem by training models on datasets with different entity types and relationships while sharing parameters. We hypothesize that the proposed approach can make more efficient use of the data and encourage the models to learn representations of words and characters (shared between multiple corpora) more effectively and generalize.

We define the multi-task setting as follows. Assume that we have $E$ datasets ($D_e$, $e \in \{1, \ldots, E\}$) for the NER task, each consisting of various entities like disease, chemicals, protein, etc. Each dataset consists of $n_e$ training texts, with length $L$. We define the loss function in the NER task $L_{\mathrm{ner}}$ as follows:
\begin{gather*}
    D_e = \{(\mathit{token}_i, y_i)_j\}, \quad \forall j \in \{1, \ldots, n_e\} \\
    L_{\mathrm{ner}} = -\sum_{e=1}^{E} \sum_{j=1}^{n_e} \sum_{i=1}^{L} \log P(\hat{y}_i)
\end{gather*}
where $P(\hat{y}_i)$ represents the model's assigned probability for the correct class $y_i$ of each token. The higher this probability, the lower the loss.

Now, for relation extraction, we have $R$ datasets ($D_r$, $r \in \{1, \ldots, R\}$), each consisting of $n_r$ sentences with entities mentioned. Each dataset has relations between two entities. We define each dataset and loss function in the RE task as follows:
\begin{gather*}
    D_r = \{(\mathit{text}, y)_j\}, \quad \forall j \in \{1, \ldots, n_r\} \\
    L_{\mathrm{re}} = -\sum_{r=1}^{R} \sum_{i=1}^{n_r} \log P(\hat{y}_i)
\end{gather*}
where $P(\hat{y}_i)$ represents the probability assigned by the model to the correct class $y_i$ for each text during classification.

Finally, the overall loss function in the multi-task setting is defined as follows:
\begin{equation*}
    L_{\mathrm{overall}} = L_{\mathrm{ner}} + L_{\mathrm{re}}
\end{equation*}

\begin{figure*}[t]
    \centering
    \includegraphics[width=\textwidth]{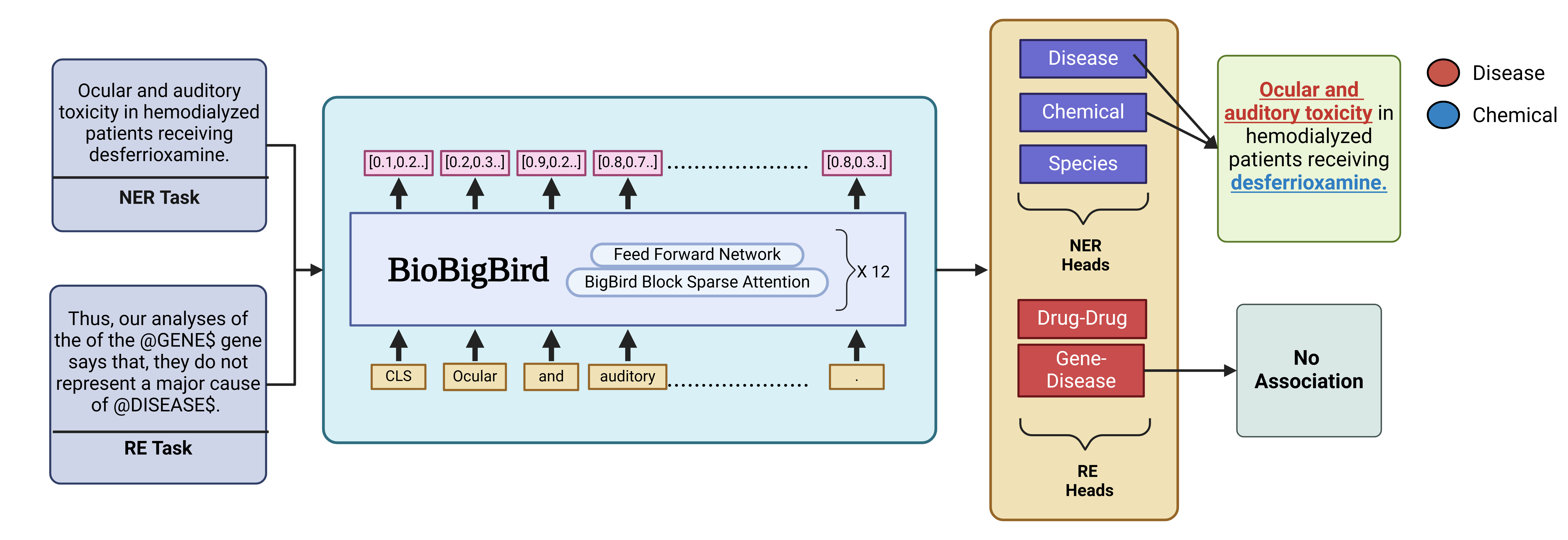}
    \caption{Overview of our multi-task architecture: we fine-tune the BioBigBird model using this method.}
    \label{fig:architecture}
\end{figure*}

The model's architecture, as illustrated in \cref{fig:architecture}, uses a transformer-based encoder. We have tried all three stages of BioBigBird to investigate its performance. Each task has its own classification head in the model. Token-level outputs are extracted for the NER task, which are subsequently directed toward a corresponding entity head for classification. For the RE task, an average pooling operation is applied to the outputs, and the pooled representations are fed into a relation extraction head depending on the entities present in the input.

\section{Experimental Setup}

\subsection{Datasets}

\Cref{tab:ner_datasets,tab:re_datasets} present the details of the datasets used to train our model, encompassing both the BLURB benchmark and some additional datasets for NER and RE tasks. In the following paragraphs, we provide descriptions of each dataset.

\begin{table}[t]
\centering
\footnotesize
\caption{Biomedical named entity recognition datasets.}
\label{tab:ner_datasets}
\begin{tabular}{@{}l@{\hspace{4pt}}l@{}}
\toprule
\textbf{Dataset} & \textbf{Entity Type} \\
\midrule
NCBI Disease \citep{dougan2014ncbi} & Disease \\
BC5CDR \citep{li2016biocreative} & Disease \\
BC5CDR \citep{li2016biocreative} & Drug/Chem. \\
BC4CHEMD \citep{krallinger2014chemdner} & Drug/Chem. \\
BC2GM \citep{smith2008overview} & Gene \\
JNLPBA \citep{collier2004introduction} & Biological Molecules \\
LINNAEUS \citep{Gerner2010} & Species \\
Species-800 \citep{pafilis2013species} & Species \\
\bottomrule
\end{tabular}
\end{table}

\begin{table}[t]
\centering
\footnotesize
\caption{Biomedical relation extraction datasets.}
\label{tab:re_datasets}
\begin{tabular}{@{}l@{\hspace{4pt}}l@{}}
\toprule
\textbf{Dataset} & \textbf{Entity Type} \\
\midrule
GAD \citep{bravo2015extraction} & Gene-disease \\
DDI \citep{herrero2013ddi} & Drug-drug \\
ChemProt \citep{krallinger2017overview} & Protein-chemical \\
\bottomrule
\end{tabular}
\end{table}

\paragraph{BC5CDR} The dataset addresses gaps in existing biomedical datasets, offering comprehensive annotations for chemical-disease entities, crucial for enhancing text-mining systems in drug discovery and biocuration. The corpus comprises 1,500 PubMed articles with 4,409 annotated chemicals, 5,818 diseases, and 3,116 chemical-disease interactions.

\paragraph{NCBI-Disease} The NCBI disease corpus, comprising 793 PubMed abstracts, offers rich disease annotations. The corpus, with 6,892 mentions, is valuable for advancing biomedical natural language processing.

\paragraph{BC4CHEMD} BC4CHEMD is a dataset consisting of 10,000 PubMed abstracts, totaling 84,355 chemical entity mentions.

\paragraph{BC2GM} BC2GM is a dataset used for the BioCreative II gene mention tagging task. It contains 20,000 sentences from abstracts of biomedical publications and is annotated with 24,583 mentions of the names of genes and related entities.

\paragraph{JNLPBA} JNLPBA is the dataset for the Joint Workshop on NLP in Biomedicine and its Application Shared task. The GENIA Project was organized based on the annotations of the GENIA Term corpus and consists of 2,404 publication abstracts. It is widely used for evaluating multi-class biomedical entity taggers.

\paragraph{LINNAEUS} The LINNAEUS corpus comprises 100 randomly selected full-text documents from the PMCOA document set. Each instance of species terms has been annotated manually in this corpus.

\paragraph{Species-800} Species-800 is a corpus for species entities. It comprises 800 PubMed abstracts that contain identified organism mentions.

\paragraph{GAD} The Gene Associations Database (GAD) compiles gene-disease associations from genetic association studies. It is a valuable resource for researchers exploring the genetic basis of diseases.

\paragraph{DDI} The DDI dataset focuses on drug-drug relations, which are crucial for patient safety and cost-effective healthcare. The DDI extraction 2013 corpus is a collection of 792 texts selected from the DrugBank database and 233 other Medline abstracts.

\paragraph{ChemProt} The ChemProt dataset comprises 1,820 PubMed abstracts featuring chemical-protein interactions annotated by domain experts, utilized in the BioCreative VI text mining shared task.

\subsection{Implementation Details}

Our BioBigBird model is based on the Flax implementation of BigBird from the HuggingFace Transformers library \citep{wolf2019huggingface}. To prepare our data, we first trained a custom text tokenizer on the PubMed articles corpus, creating a vocabulary of 32,000 tokens. The entire pre-training process was conducted for 15 days on a single TPU v2-8, during which the model was trained on a total of 22 billion tokens.

For downstream NER and RE tasks, the BioBigBird encoder generates a 768-dimensional embedding for each token. This output serves as the shared input for all task-specific heads. For NER tasks, the token embeddings are passed to a classification head consisting of a single 768-unit hidden layer with ReLU activation. For RE tasks, we first apply an average pooling operation to the token embeddings to create a single 768-dimensional representation for the sequence, which is then fed into a similar classification head.

During multi-task training, batches were constructed with a mix of examples from all NER and RE tasks. All fine-tuning experiments were conducted on three GeForce GTX 1080 Ti GPUs, using the Adam optimizer with a learning rate of $2 \times 10^{-5}$ and an epsilon of $10^{-8}$ for parameter updates.

\subsection{Evaluation}
BioBigBird's performance is evaluated on the BLURB benchmark. It is a dataset to evaluate NLP models on six tasks: 1) NER, 2) PICO annotations, 3) relation extraction, 4) document classification, 5) sentence similarity, and 6) question answering. The named entities used for model performance evaluation are chemicals (BC5-chem), diseases (BC5-disease, NCBI-disease), gene mentions (BC2GM) and terms related to molecular biology (JNLPBA). PICO annotations are annotations of texts in abstracts such as the Patient population enrolled, the Interventions studied and to what they were Compared, and the Outcomes measured (the `PICO' elements). Relation extractions used in this study are chemical-protein interaction (ChemProt), Gene-Disease association (GAD), and Drug-Disease interaction (DDI). The document classification task assigns the documents to the relevant classes. This study's categories are hallmarks of cancer taxonomy (HoC). The sentence similarity task determines the semantic similarity between the sentences (BIOSSES). Question-answering tasks evaluate the model's performance in answering the given input of the question (PubMed QA, BioASQ).

\section{Results}

\begin{table*}[t]
\centering
\caption{Comparison of performance of pre-trained models on the BLURB benchmark.}
\label{tab:blurb}
\resizebox{\textwidth}{!}{%
\begin{tabular}{lcccc}
\toprule
 & \textbf{PubMedBERT} & \textbf{BioLinkBERT} & \textbf{BioLinkBERT} & \textbf{BioBigBird} \\
 & \citep{yugu} & \citep{yasunaga2022linkbert} & \citep{yasunaga2022linkbert} & Stage 3 \\
 & Base (110 M) & Base (110 M) & Large (340 M) & (113 M) \\
\midrule
BC5-chem \citep{li2016biocreative} & 93.33 & 93.75 & 94.04 & \textbf{94.06} \\
BC5-disease \citep{li2016biocreative} & 85.62 & 86.10 & \textbf{86.39} & 85.18 \\
NCBI-disease \citep{dougan2014ncbi} & 87.82 & 88.18 & \textbf{88.76} & 85.59 \\
BC2GM \citep{smith2008overview} & 84.52 & 84.90 & \textbf{85.18} & 84.37 \\
JNLPBA \citep{collier2004introduction} & \textbf{80.06} & 79.03 & \textbf{80.06} & 78.55 \\
\midrule
EBM PICO \citep{nye2018corpus} & 73.38 & 73.97 & 74.19 & \textbf{94.6} \\
\midrule
ChemProt \citep{krallinger2017overview} & 77.24 & 77.57 & \textbf{79.98} & 74.45 \\
GAD \citep{bravo2015extraction} & 82.36 & 84.39 & 84.90 & \textbf{88.93} \\
DDI \citep{herrero2013ddi} & 82.34 & 82.72 & 83.35 & \textbf{87.33} \\
\midrule
HoC \citep{baker2016automatic} & 82.32 & 84.35 & 84.87 & \textbf{89.70} \\
\midrule
BIOSSES \citep{souganciouglu2017biosses} & 92.30 & 93.25 & \textbf{93.63} & 93.51 \\
\midrule
PubMed QA \citep{jin2019pubmedqa} & 55.84 & 70.20 & \textbf{72.18} & 59.82 \\
BioASQ \citep{nentidis2020results} & 87.56 & 91.43 & \textbf{94.82} & 84.00 \\
\midrule
BLURB Score & 81.10 & 83.39 & 84.30 & \textbf{86.46} \\
\bottomrule
\end{tabular}}
\end{table*}

\begin{table*}[t]
\centering
\caption{Comparison of performance of multi-task BioBigBird for NER and RE tasks.}
\label{tab:mt_results}
\resizebox{\textwidth}{!}{%
\begin{tabular}{lcccccc}
\toprule
 & \textbf{PubMedBERT} & \textbf{BioLinkBERT} & \textbf{BioLinkBERT} & \textbf{BERN-2} & \textbf{BioBigBird} & \textbf{BioBigBird} \\
 & \citep{yugu} & \citep{yasunaga2022linkbert} & \citep{yasunaga2022linkbert} & \citep{Sung_2022} & Stage 3 & MT-Stage 3 \\
 & Base (110 M) & Base (110 M) & Large (340 M) & (365 M) & (113 M) & (120 M) \\
\midrule
BC5-chem \citep{li2016biocreative} & 93.33 & 93.75 & 94.04 & N/A & 94.06 & \textbf{95.07} \\
BC5-disease \citep{li2016biocreative} & 85.62 & 86.10 & 86.39 & N/A & 85.18 & \textbf{90.36} \\
NCBI-disease \citep{dougan2014ncbi} & 87.82 & 88.18 & 88.76 & 88.6 & 85.59 & \textbf{93.67} \\
BC2GM \citep{smith2008overview} & 84.52 & 84.90 & 85.18 & 83.7 & 84.37 & \textbf{91.85} \\
JNLPBA \citep{collier2004introduction} & \textbf{80.06} & 79.03 & \textbf{80.06} & 78.40 & 78.55 & 77.40 \\
BC4CHEMD \citep{krallinger2014chemdner} & N/A & N/A & N/A & 92.8 & N/A & \textbf{95.01} \\
LINNAEUS \citep{Gerner2010} & N/A & N/A & N/A & \textbf{92.7} & N/A & 83.74 \\
Species-800 \citep{pafilis2013species} & N/A & N/A & N/A & N/A & N/A & 86.00 \\
\midrule
ChemProt \citep{krallinger2017overview} & 77.24 & 77.57 & 79.98 & N/A & 74.45 & \textbf{89.38} \\
GAD \citep{bravo2015extraction} & 82.36 & 84.39 & 84.90 & N/A & 88.93 & 78.08 \\
DDI \citep{herrero2013ddi} & 82.34 & 82.72 & 83.35 & N/A & 87.33 & \textbf{93.47} \\
\bottomrule
\end{tabular}}
\end{table*}

\subsection{Single-task Fine-Tuning}

The pre-trained BioBigBird model was fine-tuned on the BLURB benchmark, a widely used benchmark for evaluating biomedical language understanding tasks. To achieve optimal performance, hyperparameters were tuned separately for each downstream task. The fine-tuning process helped the model to adapt to the specific nuances of the tasks and further improve its performance.

Our approach resulted in a highly competitive model outperforming many state-of-the-art models, including the widely-used PubMedBERT and BioLinkBERT \citep{yasunaga2022linkbert} (see \cref{tab:blurb}). Specifically, our final model achieved an average F1 score of 85.55 on Named Entity Recognition (NER), an accuracy of 71.91 on question answering, a Pearson correlation of 93.51 on sentence similarity, and a Micro F1 score of 89.7 on document classification. Our approach outperforms the existing models in EBM PICO, GAD (gene-disease relationships), drug-drug interactions, and document classification. It shows that the complete article information helps improve long-range dependency.

\subsection{Multi-task Fine-Tuning}

The pre-trained BioBigBird model was fine-tuned using a multi-task framework to improve its performance in NER and RE tasks. We report the results for stage-3 BioBigBird, which performed the best, on the BLURB benchmark \citep{yugu} (see \cref{tab:mt_results}).

This approach resulted in a highly accurate model beating state-of-the-art models. The model achieves the best score on all datasets except three: JNLPBA, LINNAEUS, and GAD. The model obtained additional training data by employing multi-task fine-tuning, leading to improved generalization capabilities. Multi-task model performance is substantially higher in NER datasets like BC5-Chem, BC5-disease, NCBI-disease, BC2GM and RE datasets like ChemProt and DDI. The multi-task system performs better than single-task fine-tuning on most datasets, making it a promising approach for enhanced fine-tuning.

\section{Conclusions}

We propose using BigBird, a transformer-based language model, for natural language processing tasks in the biomedical domain. BigBird is designed to efficiently handle long input sequences using a block sparse attention mechanism, which reduces the quadratic dependency on sequence length to linear. To evaluate the performance of BigBird on these tasks, we pre-train the model on a large dataset of biomedical articles, similar to PubMedBERT, but using the full articles rather than just the abstracts. Pre-training the model on a large dataset of full biomedical articles, rather than just the abstracts, can lead to an improvement in performance because of providing more context for the model through longer sequences during training. Overall, BigBird has the potential to be a valuable tool for researchers in the biomedical field. Its ability to effectively process long sequences and achieve superior performance on a range of natural language processing tasks makes it an attractive choice for various applications in this domain.

We also introduce a multi-task approach for fine-tuning NER and RE tasks. This approach maximizes the utilization of annotated data within the domain. The shared representation learning enhances the overall model performance. Additionally, it offers a unified model capable of addressing different NER and RE tasks, eliminating the cumbersome process of training and saving a model for each task.

\section*{Limitations}
The usage of PubMed articles for pre-training, although rich in information content, may introduce noise in the fine-tuning stage where the typical sequence length is below 1024.

\section*{Ethical Considerations}
BioBigBird is trained on the same corpora as existing language models. The model might reflect toxic behaviour and biases such as race, gender and demography.

\bibliography{custom}

\end{document}